\documentclass[conference,a4paper]{IEEEtran}
\IEEEoverridecommandlockouts

\usepackage[pagebackref,breaklinks,colorlinks]{hyperref}
\usepackage[cmex10]{amsmath}%American Math Society(AMS) math formatting
\usepackage{amssymb,amsfonts}%AMS extra symbols and fonts
\usepackage[ruled,vlined]{algorithm2e}
\usepackage{graphicx}
\graphicspath{{Figures/PDF/}{Figures/PNG/}}

\usepackage{booktabs}
\usepackage{siunitx}
\usepackage[numbers,compress]{natbib}
\usepackage{bm,bbm}

\usepackage{multirow}
\usepackage{tabularx}

\usepackage[a4paper, left=19mm, right=19mm, top=25mm, bottom=25mm]{geometry}

\begin{document}

\title{
    \uppercase{
        A Simple Aerial Detection Baseline \\
        of Multimodal Language Models\\
    }
}

\author{
    \IEEEauthorblockN{Qingyun Li$^{1}$, Yushi Chen$^{1\dagger}$, Xinya Shu$^{1}$, Dong Chen$^{1}$, Xin He$^{1}$, Yi Yu$^{2}$, Xue Yang$^{3\ddagger}$}
    \IEEEauthorblockA{
        \textit{$^1$Harbin Institute of Technology}~
        \textit{$^2$Southeast University}~
        \textit{$^3$Shanghai Jiao Tong University}
    }
    \IEEEauthorblockA{$^{\dagger}$ Corresponding author \quad $^{\ddagger}$ Project leader}
    \IEEEauthorblockA{Code: \url{https://github.com/Li-Qingyun/mllm-mmrotate}}
}

\maketitle
\vspace{-20pt}

\begin{abstract}
The multimodal language models (MLMs) based on generative pre-trained Transformer are considered powerful candidates for unifying various domains and tasks. MLMs developed for remote sensing (RS) have demonstrated outstanding performance in multiple tasks, such as visual question answering and visual grounding.
In addition to visual grounding that detects specific objects corresponded to given instruction, aerial detection, which detects all objects of multiple categories, is also a valuable and challenging task for RS foundation models. However, aerial detection has not been explored by existing RS MLMs because the autoregressive prediction mechanism of MLMs differs significantly from the detection outputs.
In this paper, we present a simple baseline for applying MLMs to aerial detection for the first time, named LMMRotate. Specifically, we first introduce a normalization method to transform detection outputs into textual outputs to be compatible with the MLM framework. Then, we propose a evaluation method, which ensures a fair comparison between MLMs and conventional object detection models. We construct the baseline by fine-tuning open-source general-purpose MLMs and achieve impressive detection performance comparable to conventional detector. We hope that this baseline will serve as a reference for future MLM development, enabling more comprehensive capabilities for understanding RS images. 
\end{abstract}


\section{Introduction}

Earth observation systems have acquired vast remote sensing (RS) data, driving demand for automated RS image interpretation. The advancement of artificial general intelligence (AGI) has motivated researchers in this field to develop general agents that are outstanding on multiple tasks, such as scene classification, visual question answering, and object detection~\cite{zhou2024vlgfm,kuckreja2023geochat,zhang2024earthgpt,luo2024skysensegpt}.

Multimodal language models (MLMs) are built upon vision and language foundation models, enabling them to process data from multiple modalities and interpret textual instructions effectively.
The task results are outputted in textual form. By leveraging powerful pre-trained foundation models and a flexible text interface, MLMs are considered a key component in the advancement of AGI.

\begin{figure}
    \centering
    \includegraphics[width=\linewidth]{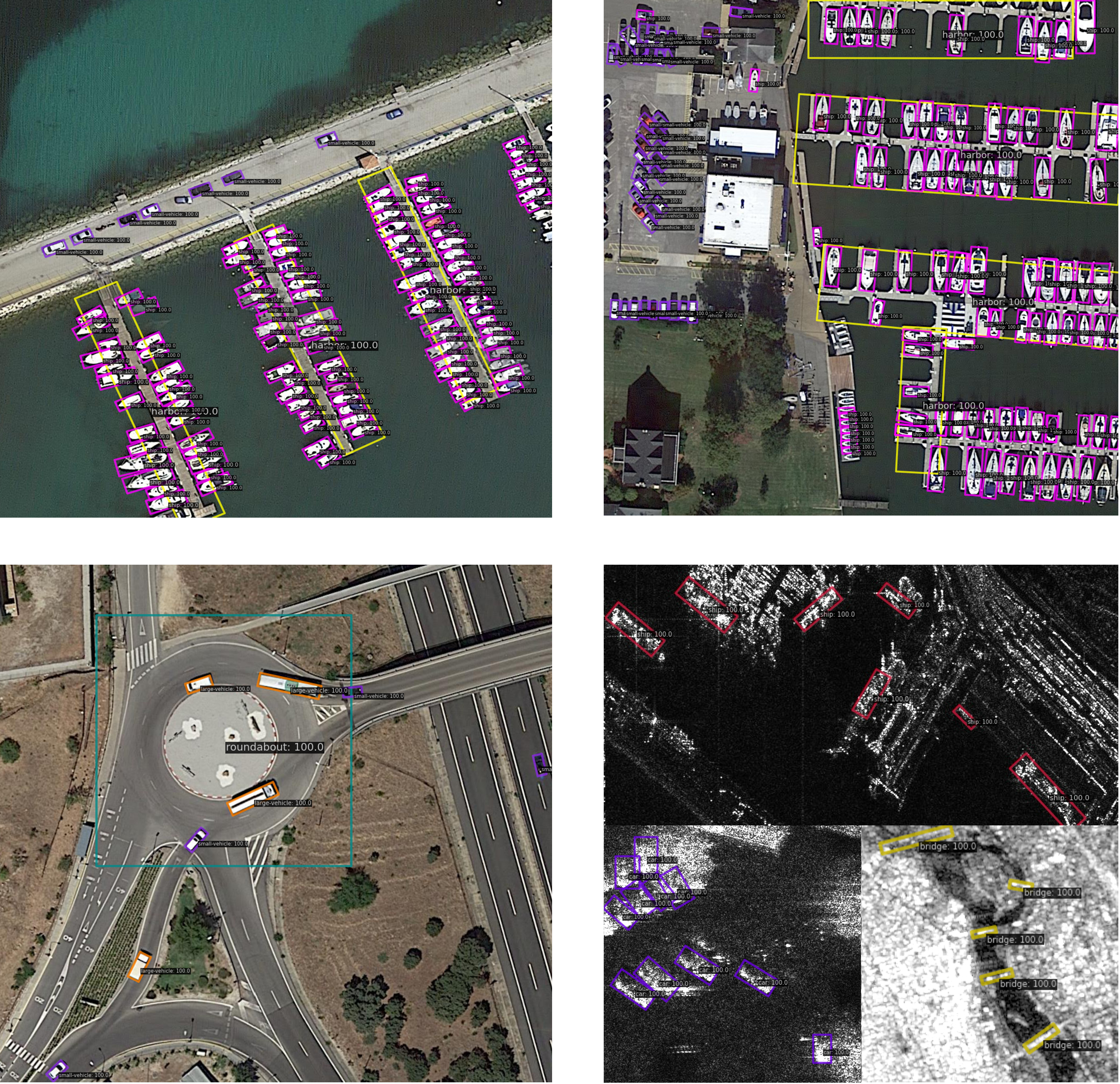}
    \caption{Visualization of the objects detected by our MLM detector based on Florence-2-large~\cite{florence2} with single dataset setting. The images are selected from the test sets of DOTA-v1.0~\cite{DOTA} and RSAR~\cite{RSAR}.}
    \label{fig:visualization}
    \vspace{-4mm}
\end{figure}

Recently, MLMs have been introduced into the RS field. While existing RS MLMs have emphasized visual localization capabilities, their detection performance still requires significant improvement.
Geochat~\citep{kuckreja2023geochat} presents the first open-source RS MLM learned on tasks including scene classification, image captioning, region description, visual question answering, and visual grounding. The proposed model demonstrates the ability to detect partial objects corresponding to textual instructions but is unable to perform more intensive detection.
EarthGPT~\citep{zhang2024earthgpt} extends visual inputs with multisensor RS modalities covering optical, synthetic aperture radar, infrared data. While they achieve detection results comparable to traditional detectors, their model is limited to single-class detection performance.
SkySenseGPT~\citep{luo2024skysensegpt} introduces scene graph generation to RS MLMs to enhance ability of understanding relations between objects. They also provide tolerance detection accuracies under a lower threshold of intersection over union (IoU).
We consider that RS MLMs still require further investigation in aerial detection to fully explore their potential.

There is often skepticism about whether MLMs can effectively learn to perform aerial detection. First, the detection outputs consist of numerical coordinates for bounding boxes and object categories, which significantly differ from the textual outputs produced by language models. Second, language generation models are typically autoregressive, generating causal sequences, whereas detection models usually output all results in parallel. Additionally, aerial detection presents considerable challenges due to the presence of many small and densely packed objects, which impose high demands on both the visual input resolution and the output sequence length of MLMs.

In this paper, we present a simple baseline for aerial detection, with a focus on enhancing the detection capabilities of MLMs in the RS domain for the first time, named LMMRotate. Specifically, we propose a straightforward method to supervised fine-tune MLMs, achieving detection performance comparable to that of conventional detectors. The MLM detectors produce parsable text outputs, offering both flexibility and expandability as multi-dataset joint-trained detectors. We then observe that the advantage of conventional detectors in detection metrics largely comes from object confidence scores. To address this, we propose an appropriate evaluation scheme that enables fair comparisons between MLMs and conventional detectors. Additionally, we conduct exploratory experiments using open-source MLMs, demonstrating the potential of MLMs as aerial detectors. We sincerely hope that our work contributes to advancing more comprehensive abilities of MLMs in the RS field.

\section{Method}

\subsection{Preliminary of Multimodal Language Models}

\begin{figure}
    \centering
    \includegraphics[width=0.9\linewidth]{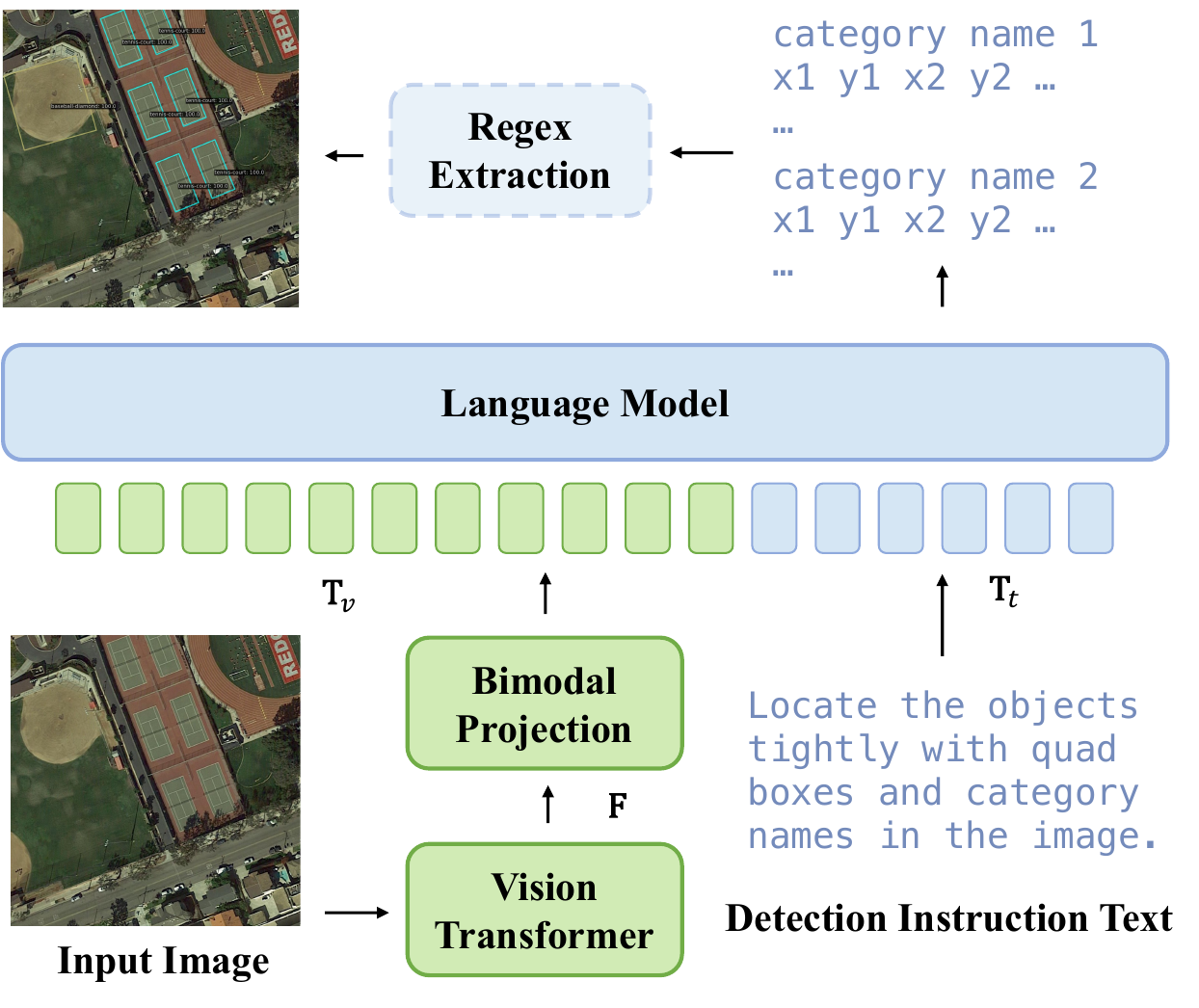}
    \caption{The overall framework of the proposed MLM detector baseline.}
    \label{fig:framework}
    \vspace{-2mm}
\end{figure}

As depicted in Figure~\ref{fig:framework}, the proposed method adopts the most popular MLM paradigm for image understanding, which connects a vision foundation model and a language foundation model using a bimodal projection operation. We fine-tune off-the-shelf pre-trained MLMs to inherit the localization abilities learned from grounding task. 

For a language model, the input sentence is pieced and tokenized into a unique sequence of vocabulary indices. Each indice $i$ corresponds to a discrete token $\mathbf{t}_{i} \in \mathbb{R}^{D}$, which is a learnable vector of dimensionality $D$. The model's output is also a sequence of indices, which is subsequently de-tokenized to generate the final response.

An input RS image is firstly transformed into flattened features $\mathbf{F} \in \mathbb{R}^{N_I \times D_I}$ by image preprocessing operations (resizing or dynamic resolution strategies) and a well-designed vision Transformer, where $N_I$ and $D_I$ represent the number and dimensionality of the features, respectively. Meanwhile, the prompt text of detection instruction is tokenized into $N_t$ text tokens $\mathbf{T}_t \in \mathbb{R}^{N_t \times D}$. To align the visual features to input space of the language model, a bimodal projection operation maps the image features into $N_v$ visual tokens $\mathbf{T}_v \in \mathbb{R}^{N_v \times D}, N_v \propto N_I$. The input of language model is:
\begin{equation}
\mathbf{T} = \texttt{concat} ( \mathbf{T}_v,\mathbf{T}_t ) \in \mathbb{R}^{ ( N_v + N_t ) \times D},
\end{equation}
where the $\texttt{concat}(~,~)$ operation concatenates two matrices along the token number dimension.

In the training period, we optimize the model parameters $\theta$ using the standard language modeling strategy, i.e., next token prediction with cross-entropy loss:
\begin{equation}
\mathcal{L} = -  \sum_{j=1}^{|\textbf{r}|}P_j(\mathbf{r},\mathbf{T}), ~~ P_j(\mathbf{r},\mathbf{T}) = logP_{\theta}(\textbf{r}_j|\textbf{r}_{<j}, \mathbf{T}),
\end{equation}
where $\mathbf{r} = (\mathbf{r}_1, \mathbf{r}_2, \dots, \mathbf{r}_T)$ denotes the indices sequence of model response and $P_j(\mathbf{r},\mathbf{T})$ is the conditional probability distribution of the $j$-th token. During the inference phase, the model generates outputs in an auto-regressive manner, predicting tokens iteratively, . The $j$-th token is obtained:
\begin{equation}
\mathbf{r}_j = \arg\max P_j(\mathbf{r},\mathbf{T}) ~~ \text{or} ~~ \mathbf{r}_j \sim P_j(\mathbf{r},\mathbf{T}),
\end{equation}
where the former corresponds to deterministic methods, such as greedy search or beam search, while the latter corresponds to stochastic sampling strategies.
The process continues until a stopping criterion is met, such as generating the end-of-sequence token.

\vspace{-1pt}
\subsection{Normalization of Detection Outputs}

\begin{figure}[t]
    \centering
    \includegraphics[width=\linewidth]{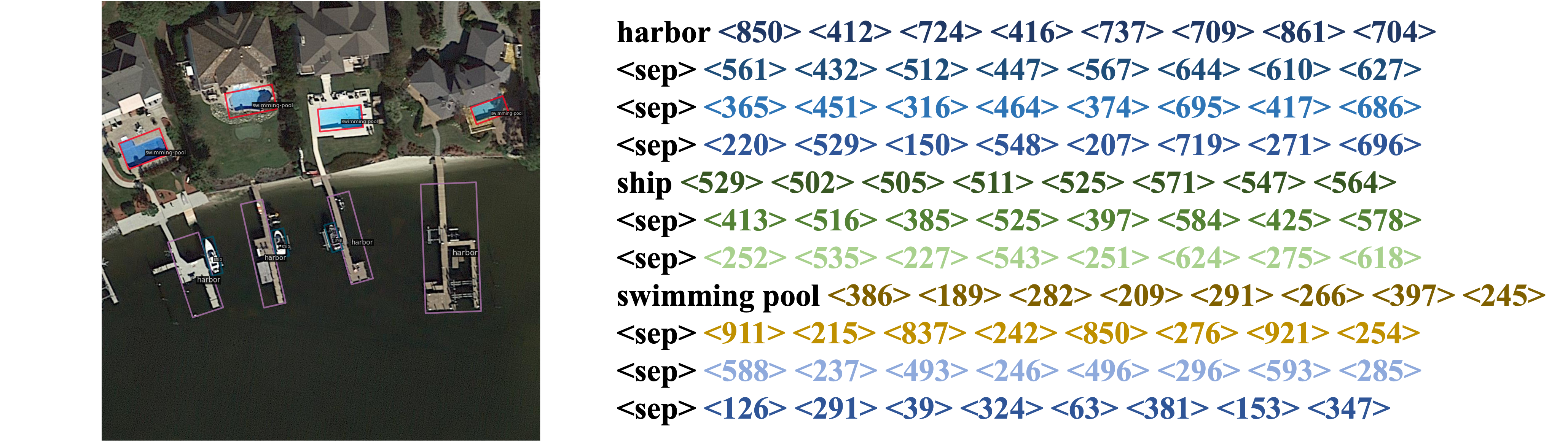}
    \caption{An example of a RS image and its response that contains category names and 8-parameter polygon boxes of objects.}
    \label{fig:response_example}
    \vspace{-2mm}
\end{figure}

We investigate multi-class orientated aerial detection in this paper. Each object is represented with the class name and a 8-parameters quadrilateral bounding box $\mathbf{o}=(n_\mathbf{o}, x_{1\mathbf{o}}, y_{1\mathbf{o}}, x_{2\mathbf{o}}, y_{2\mathbf{o}}, x_{3\mathbf{o}}, y_{3\mathbf{o}}, x_{4\mathbf{o}}, y_{4\mathbf{o}})$, where the $x_{i\mathbf{o}}$ and $y_{i\mathbf{o}}$ are the coordinates of the polygon vertices in clockwise order. The vertex with the smallest vertical coordinate is considered the starting vertex. Besides, $n_{\mathbf{o}}$ is one phrase in the $c$ proposal category names $\{C_1, C_2, ..., C_c\}$.

We follow the approaches in Florence-2~\citep{florence2} and InternVL~\citep{internvl_2_5} to qualitize vertex coordinates by normalizing each axis from 0 to 1000 and then round the normalized coordinate into integer. This designing prevents the coordinates from being excessively continuous and ensures consistent predictions. The precision loss introduced by the quantized integer coordinates relative to the floating-point coordinates has a negligible impact on the location accuracy. 

The template used to standardize detection annotations should ensure uniqueness and orderliness. For a given image, the model should output the detected objects in a logical and sequential manner. Specifically, the overall response is composed of detection results for each category, which are sorted alphabetically by category names. Within each category, the boxes are further sorted based on the position of the starting vertexes. 

Figure~\ref{fig:response_example} exhibits an example RS image and its corresponding response for Florence-2 model, where ``$\langle v \rangle$'' and ``$\langle sep \rangle$'' mean the $v$-th bin of the corresponding axis and separating marker, respectively. Additionally, categories are directly represented in textual form. 

As shown, we enable the MLM to recognize objects of multiple categories in an aerial image, including categories and bounding box in the response. During the inference stage, the detection results can be extracted through simple regex processing from the response. Moreover, compared to most conventional detectors that rely on post-processing techniques like non-maximum suppression (NMS), MLM does not need to handle issues related to overlapping redundant objects.

\vspace{-2pt}
\subsection{Evaluation of MLM detectors}

\label{sec:method-eval}

The common practices for aerial detection use mean average precision (mAP) as evaluation metric, which requires orientated bounding boxes, categories, and confidence scores of the detected objects. As mentioned, the designed response only includes categories and location. We found that the quality of the confidence scores has a significant impact on the mAP score, which is a disadvantage for MLM.

\begin{figure}[t]
    \centering
    \includegraphics[width=\linewidth]{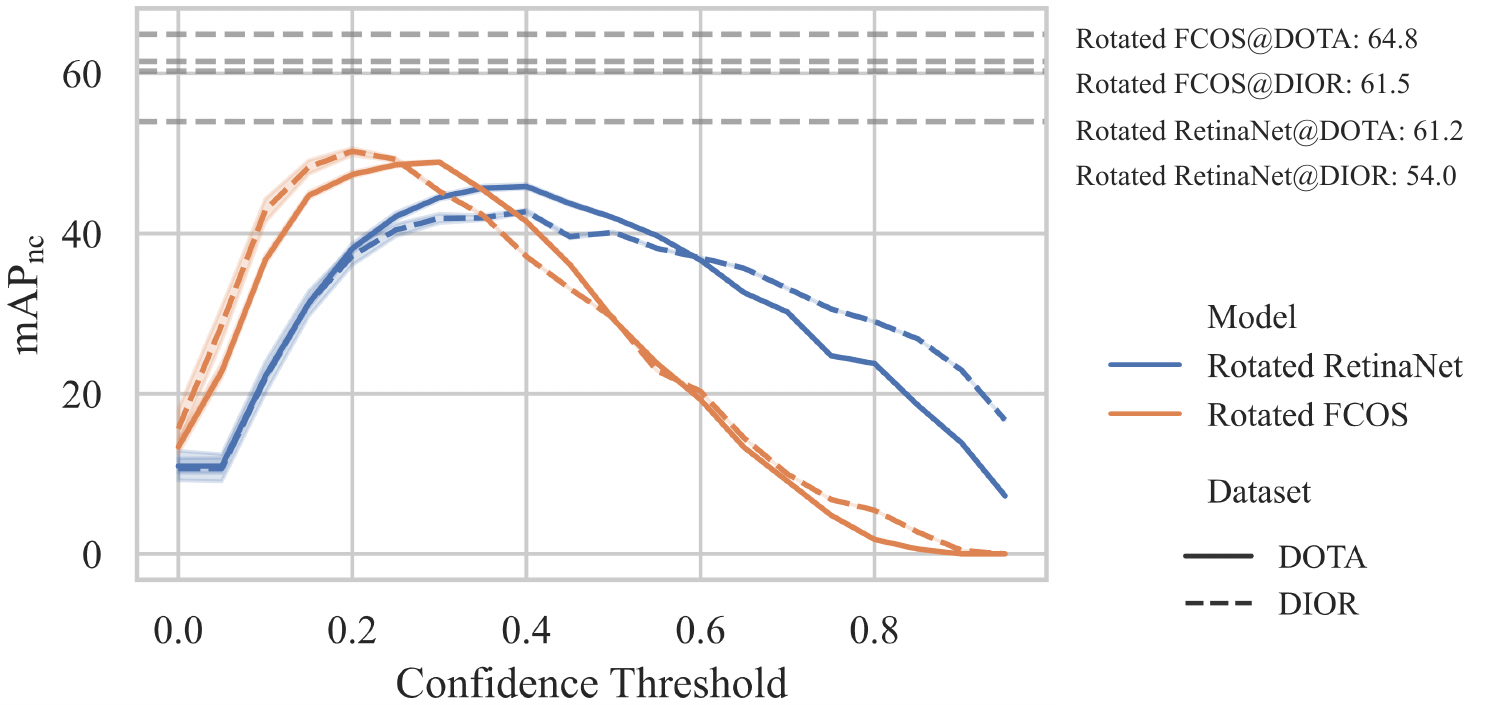}
    \caption{The impact of confidence scores on mAP / $\text{mAP}_{\text{nc}}$ with error bands. The colored lines record the variation trends of $\text{mAP}_{\text{nc}}$ for the two popular conventional detector on DOTA-v1.0~\citep{DOTA} (trained on `train` split and evaluated on `validation` split) and DIOR-R~\citep{DIOR} (trained on `trainval` split and evaluated on `test` split, and the input size is $800 \times 800$) datasets under different confidence thresholds.}
    \label{fig:score_impact}
    \vspace{-2mm}
\end{figure}

Figure~\ref{fig:score_impact} demonstrates the impact of confidence scores. The gray line represents the original mAP, while the colored line represents the mAP with confidence substituted with constant or random scores, referred to as mAP with no confidence, or $\text{mAP}_{\text{nc}}$. Although the predictions of conventional detectors have undergone NMS, the directly obtained $\text{mAP}_{\text{nc}}$ is low due to low-confidence false positive boxes. Hence, we conduct threshold filtering to enhance the $\text{mAP}_{\text{nc}}$ and set the best result as the $\text{mAP}_{\text{nc}}$ of conventional detectors. It can be observed that the presence of confidence leads to a significant increase on mAP.

Rather than introducing a complex mechanism to obtain object confidence for MLM, we claim that the consideration of confidence is not necessary when comparing MLM and conventional detectors. Since the annotations and results for detection tasks inherently compose of object categories and bounding boxes, but not contains confidence. The confidence is merely an additional byproduct of the detector’s inference process. It can assist in processing detection results but is not essential for evaluating model quality. We advocate for the use of metrics that do not rely on confidence, such as $\text{mF}_\text{1}$ and $\text{mAP}_{\text{nc}}$. 

We also evaluate the robustness of $\text{mAP}_{\text{nc}}$ as a metric. We calculate each $\text{mAP}_{\text{nc}}$ eleven times by replacing the confidence with ten random values and a consistent value. As shown in Figure~\ref{fig:score_impact}, the standard deviations are generally lower than 0.5\%, especially when the threshold is between 0.2 and 0.4.

Especially for benchmarks that lack publicly available test sets and require online server evaluation based on mAP, such as DOTA~\citep{DOTA} and FAIR1M~\citep{FAIR1M}, hence, we recommend adopting $\text{mAP}_{\text{nc}}$ as the evaluation metric in these benchmarks.

\section{Experiment}

\begin{table*}[!tb]
\centering
\caption{Comparison Results of Two Conventional Detectors and Our MLM Detectors on Four Benchmarks}
\label{tab:results}
\setlength{\tabcolsep}{2.0mm}
\renewcommand{\arraystretch}{1.2} 
\begin{tabular}{@{}l|cc|cc|cc|cc|cc@{}}
\toprule
\multicolumn{1}{c|}{\multirow{2}{*}{\textbf{Model}}} & \multicolumn{2}{c|}{\textbf{DOTA-v1.0}~\citep{DOTA}} & \multicolumn{2}{c|}{\textbf{DIOR-R}~\citep{DIOR}} & \multicolumn{2}{c|}{\textbf{FAIR1M-v1.0}~\citep{FAIR1M}} & \multicolumn{2}{c|}{\textbf{SRSDD}~\citep{SRSDD}} & \multicolumn{2}{c}{\textbf{RSAR}~\citep{RSAR}} \\ \cline{2-11} 
\multicolumn{1}{c|}{} & $\text{mAP}_{\text{nc}}$ & $\text{mF}_\text{1}$ & $\text{mAP}_{\text{nc}}$ & $\text{mF}_\text{1}$ & $\text{mAP}_{\text{nc}}$ & $\text{mF}_\text{1}$ & $\text{mAP}_{\text{nc}}$ & $\text{mF}_\text{1}$ & $\text{mAP}_{\text{nc}}$ & $\text{mF}_\text{1}$ \\ \hline
Rotated RetinaNet~\citep{retinanet} & 52.2 & 58.2 & 43.3 & 56.3 & 21.4 & 36.2 & 14.4 & 18.8 & 49.1 & 57.4 \\
Rotated FCOS~\citep{fcos} & 58.9 & 61.1 & 50.0 & 63.6 & 25.6 & 39.4 & 34.5 & 38.5 & 59.0 & 62.9 \\ \hline
Florence-2-base~\citep{florence2} & 50.2 & 62.2 & 53.0 & 66.3 & 23.6 & 40.2 & 15.3 & 25.4 & 58.8 & 64.0 \\
Joint training (concat) & 51.3 & - & 53.5 & 65.9 & 22.0 & 38.1 & 17.8 & 23.7 & 58.4 & 63.7 \\
Joint training (balanced) & 52.0 & - & 53.8 & 66.3 & 21.9 & 38.6 & 25.5 & 30.9 & 56.2 & 61.2 \\ \hline
Florence-2-large~\citep{florence2} & 56.0 & 63.1 & 54.9 & 67.9 & 26.8 & 44.0 & 10.4 & 18.2 & 63.0 & 66.5 \\
Joint training (concat) & 55.3 & - & 56.9 & 68.9 & 25.6 & 42.5 & 24.6 & 32.8 & 64.1 & 67.0 \\
Joint training (balanced) & 56.1 & - & 56.3 & 69.5 & 25.2 & 41.8 & 31.4 & 37.4 & 60.8 & 65.8 \\ \bottomrule

\end{tabular}
% \vspace{-4mm}
\vspace{-2mm}
\end{table*}

\subsection{Benchmark Datasets}

We conduct experiments on four benchmarks for multi-class orientated aerial detection, comprising three optical RS image datasets DOTA-v1.0~\citep{DOTA}, DIOR-R~\citep{DIOR}, FAIR1M-v1.0~\citep{FAIR1M}, and two synthetic aperture radar (SAR) image datasets SRSDD~\citep{SRSDD} and RSAR~\citep{RSAR}. 

\vspace{-2pt}
\subsection{Evaluation Settings}

As claimed in Section~\ref{sec:method-eval}, we use $\text{mAP}_{\text{nc}}$ and $\text{mF}_{1}$ as evaluation metrics. 

The $\text{mAP}_{\text{nc}}$ and $\text{mF}_{1}$ of each conventional detector is calculated by the process exhibited in Figure~\ref{fig:score_impact}. We calculate the two types of scores under a range of confidence thresholds and then select the best scores. We obtain the results with the default inference settings of each model and do not post-process any model with an additional NMS operation.

For the MLM detectors, we first extract the detection outputs with regular expressions. For cases in which the predicted category text does not exactly match the meta category names of the dataset (e.g., ``pool'' and ``swimming pool''), we employ fuzzy matching with the Levenshtein distance. We assign a 100\% confidence score to all the predictions. Then, we directly send the predictions to subsequent evaluation.

We emphasize a critical distinction between $\text{mAP}_{\text{nc}}$ and $\text{mF}_{1}$ is that the $\text{mF}_{1}$ cannot be utilized for online evaluation. Consequently, for DOTA-v1.0 and FAIR1M-v1.0, $\text{mF}_{1}$ can not be calculate in the commonly adopted benchmark settings. Therefore, we adjust the benchmark settings for $\text{mF}_{\text{1}}$. 
For DOTA-v1.0, 
we re-train the model on the training set and compute $\text{mF}_{1}$ on the validation set. For FAIR1M-v1.0, we directly evaluate the model on the validation set of FAIR1M-v2.0, because the newer version adds a validation set and expands the test set without modifying the training set.

\vspace{-2pt}
\subsection{Implementation Details}

We fine-tune Florence-2~\citep{florence2}, an advanced MLM for general vision tasks. We conduct experiment on the two available pre-trained model: Florence-2-base with 271 million parameters and Florence-2-large with 829 million parameters. The input size is $1024 \times 1024$ for all models.

Since our data annotations are stored in pure text format, we can flexibly perform joint training across multiple datasets. We explore two methods for merging multiple dataset: The ``concat'' method simply merges the four datasets. The ``balanced'' method oversamples the smaller datasets, thereby achieving a more balanced distribution among the four datasets.

The models are trained for 100 epochs, with the learning rate of $2 \times 10^{-5}$ and cosine schedule. We open-source the code based on MMRotate~\citep{zhou2022mmrotate} and Huggingface Transformers, which employs resource-friendly MLM training techniques, such as mixed precision computing, gradient checkpointing, flash attention, and deepspeed.

\vspace{-2pt}
\subsection{Comparison Results}

Table~\ref{tab:results} presents the comparison results of two conventional detectors and our MLM detectors on the four benchmarks. Figure~\ref{fig:visualization} exhibits the visualization of the objects detected by our MLM detector based on Florence-2-large~\citep{florence2} with single dataset setting.

Overall, the fine-tuned MLM detectors achieve detection performance on par with conventional detectors. As shown in Figure~\ref{fig:visualization}, even in complex scenarios with numerous densely packed small objects, the MLM detector can still perform well. In the quantitative results presented in Table~\ref{tab:results}, the MLM detectors even surpasses conventional detectors in terms of $\text{mF}_1$, as well as $\text{mAP}_{\text{nc}}$ on the DIOR-R, FAIR1M-v1.0, and RSAR datasets.

It can also be observed that the joint training have great positive effect on MLM detectors. The overall performance is increased because of the larger amount of data, especally for the small dataset SRSDD, which seems to gain more benifits from the other datasets.

\section{Conclusion}

In this paper, we demonstrate that a multimodal language model can also handle aerial detection tasks, even achieving performance comparable to conventional detectors. Our approach is straightforward: by normalizing detection outputs into text form, we can fine-tune general-purpose MLMs to meet our goal. We also present an appropriate evaluation scheme for MLM detectors. We hope our work will inspire future research toward enhancing aerial detection capabilities in next-generation RS MLMs, ultimately contributing to broader AGI developments.

\small
\bibliographystyle{IEEEtranN}
\bibliography{references}

% Generated by IEEEtranN.bst, version: 1.14 (2015/08/26)
\begin{thebibliography}{14}
\providecommand{\natexlab}[1]{#1}
\providecommand{\url}[1]{#1}
\csname url@samestyle\endcsname
\providecommand{\newblock}{\relax}
\providecommand{\bibinfo}[2]{#2}
\providecommand{\BIBentrySTDinterwordspacing}{\spaceskip=0pt\relax}
\providecommand{\BIBentryALTinterwordstretchfactor}{4}
\providecommand{\BIBentryALTinterwordspacing}{\spaceskip=\fontdimen2\font plus
\BIBentryALTinterwordstretchfactor\fontdimen3\font minus \fontdimen4\font\relax}
\providecommand{\BIBforeignlanguage}[2]{{%
\expandafter\ifx\csname l@#1\endcsname\relax
\typeout{** WARNING: IEEEtranN.bst: No hyphenation pattern has been}%
\typeout{** loaded for the language `#1'. Using the pattern for}%
\typeout{** the default language instead.}%
\else
\language=\csname l@#1\endcsname
\fi
#2}}
\providecommand{\BIBdecl}{\relax}
\BIBdecl

\bibitem[Zhou et~al.(2024)Zhou, Feng, Ke, et~al.]{zhou2024vlgfm}
Y.~Zhou, L.~Feng, Y.~Ke \emph{et~al.}, ``Towards vision-language geo-foundation models: A survey,'' \emph{arXiv preprint arXiv:2406.09385}, 2024.

\bibitem[Kuckreja et~al.(2024)Kuckreja, Danish, Naseer, et~al.]{kuckreja2023geochat}
K.~Kuckreja, M.~S. Danish, M.~Naseer \emph{et~al.}, ``Geochat: Grounded large vision-language model for remote sensing,'' \emph{CVPR}, 2024.

\bibitem[Zhang et~al.(2024)Zhang, Cai, Zhang, et~al.]{zhang2024earthgpt}
W.~Zhang, M.~Cai, T.~Zhang \emph{et~al.}, ``Earthgpt: A universal multimodal large language model for multisensor image comprehension in remote sensing domain,'' \emph{IEEE Trans. Geosci. Remote Sens.}, vol.~62, pp. 1--20, 2024.

\bibitem[Luo et~al.(2024)]{luo2024skysensegpt}
J.~Luo \emph{et~al.}, ``Skysensegpt: A fine-grained instruction tuning dataset and model for remote sensing vision-language understanding,'' \emph{arXiv preprint arXiv:2406.10100}, 2024.

\bibitem[Xiao et~al.(2024)Xiao, Wu, Xu, et~al.]{florence2}
B.~Xiao, H.~Wu, W.~Xu \emph{et~al.}, ``Florence-2: Advancing a unified representation for a variety of vision tasks,'' in \emph{CVPR}, June 2024, pp. 4818--4829.

\bibitem[Xia et~al.(2018)Xia, Bai, Ding, et~al.]{DOTA}
G.-S. Xia, X.~Bai, J.~Ding \emph{et~al.}, ``Dota: A large-scale dataset for object detection in aerial images,'' in \emph{CVPR}, June 2018.

\bibitem[Zhang et~al.(2025)Zhang, Yang, Li, et~al.]{RSAR}
X.~Zhang, X.~Yang, Y.~Li \emph{et~al.}, ``Rsar: Restricted state angle resolver and rotated sar benchmark,'' \emph{arXiv preprint arXiv:2501.04440}, 2025.

\bibitem[Chen et~al.(2024)Chen, Wang, Cao, et~al.]{internvl_2_5}
Z.~Chen, W.~Wang, Y.~Cao \emph{et~al.}, ``Expanding performance boundaries of open-source multimodal models with model, data, and test-time scaling,'' \emph{arXiv preprint arXiv:2412.05271}, 2024.

\bibitem[Sun et~al.(2022)Sun, Wang, Yan, et~al.]{FAIR1M}
X.~Sun, P.~Wang, Z.~Yan \emph{et~al.}, ``Fair1m: A benchmark dataset for fine-grained object recognition in high-resolution remote sensing imagery,'' \emph{ISPRS J. Photogram. Remote Sens.}, vol. 184, pp. 116--130, 2022.

\bibitem[Li et~al.(2020)]{DIOR}
K.~Li \emph{et~al.}, ``Object detection in optical remote sensing images: A survey and a new benchmark,'' \emph{ISPRS J. Photogram. Remote Sens.}, vol. 159, pp. 296--307, 2020.

\bibitem[Lei et~al.(2021)Lei, Lu, Qiu, and Ding]{SRSDD}
S.~Lei, D.~Lu, X.~Qiu, and C.~Ding, ``Srsdd-v1. 0: A high-resolution sar rotation ship detection dataset,'' \emph{Remote Sensing}, vol.~13, no.~24, p. 5104, 2021.

\bibitem[Ross and Doll{\'a}r(2017)]{retinanet}
T.-Y. Ross and G.~Doll{\'a}r, ``Focal loss for dense object detection,'' in \emph{CVPR}, 2017, pp. 2980--2988.

\bibitem[Tian et~al.(2020)Tian, Shen, Chen, and He]{fcos}
Z.~Tian, C.~Shen, H.~Chen, and T.~He, ``Fcos: A simple and strong anchor-free object detector,'' \emph{IEEE Trans. Pattern Anal. Mach. Intell.}, vol.~44, no.~4, pp. 1922--1933, 2020.

\bibitem[Zhou et~al.(2022)Zhou, Yang, Zhang, et~al.]{zhou2022mmrotate}
Y.~Zhou, X.~Yang, G.~Zhang \emph{et~al.}, ``Mmrotate: A rotated object detection benchmark using pytorch,'' in \emph{ACM MM}, 2022, pp. 7331--7334.

\end{thebibliography}

\end{document}